\documentclass{article}

\usepackage{microtype}
\usepackage{graphicx}
\usepackage{booktabs} 

\usepackage{hyperref}

\usepackage[accepted]{icml2023}

\usepackage{amsmath}
\usepackage{amssymb}
\usepackage{mathtools}
\usepackage{amsthm}
\usepackage{bibentry}

\usepackage{float}
\usepackage{caption}
\usepackage{subcaption}
\usepackage{xcolor}
\usepackage[symbol]{footmisc}
\usepackage{multirow}

\usepackage[capitalize,noabbrev]{cleveref}

\theoremstyle{plain}

\theoremstyle{definition}

\theoremstyle{remark}

\usepackage[textsize=tiny]{todonotes}

\newenvironment{alphafootnotes}
  {\par\edef\savedfootnotenumber{\number\value{footnote}}
   
   \setcounter{footnote}{0}}
  {\par\setcounter{footnote}{\savedfootnotenumber}}

\icmltitlerunning{\illume: Rationalizing Vision-Language Models through Human Interactions}
\newcommand{\illume}{\mbox{\textsc{Illume}}}

\begin{document}

\twocolumn[
\icmltitle{\illume: Rationalizing Vision-Language Models through Human Interactions}



\icmlsetsymbol{equal}{*}

\begin{icmlauthorlist}
\icmlauthor{Manuel Brack}{equal,dfki,tud}
\icmlauthor{Patrick Schramowski}{equal,dfki,tud,hai,lai}
\icmlauthor{Björn Deiseroth}{tud,hai,alp}
\icmlauthor{Kristian Kersting}{dfki,tud,hai,cog}

\end{icmlauthorlist}

\icmlaffiliation{dfki}{German Center for Artificial Intelligence (DFKI)}
\icmlaffiliation{tud}{Computer Science Department, TU Darmstadt}
\icmlaffiliation{hai}{Hessian Center for AI (hessian.AI)}
\icmlaffiliation{alp}{Aleph Alpha}
\icmlaffiliation{cog}{Centre for Cognitive Science, TU Darmstadt}
\icmlaffiliation{lai}{LAION}

\icmlcorrespondingauthor{}{\{brack,schramowski\}@cs.tu-darmstadt.de}

\icmlkeywords{Machine Learning, ICML}

\vskip 0.3in
]



\printAffiliationsAndNotice{\icmlEqualContribution} 
\begin{alphafootnotes}
\begin{abstract}

Bootstrapping from pre-trained language models has been proven to be an efficient approach for building vision-language models (VLM) for tasks such as image captioning or visual question answering. 
However, outputs of these models rarely align with user's rationales for specific answers.
In order to improve this alignment and reinforce commonsense reasons, we propose a tuning paradigm based on human interactions with machine generated data. 
Our \illume~executes the following loop: Given an image-question-answer prompt, the VLM samples multiple candidate rationales, and a human critic provides feedback via preference selection, used for fine-tuning.
This loop increases the training data and gradually carves out the VLM's rationalization capabilities that are aligned with human intent.
Our exhaustive experiments demonstrate that \illume~is competitive with standard supervised fine-tuning while using significantly fewer training data and only requiring minimal feedback.\footnote{Code available at:\\ \url{https://github.com/ml-research/ILLUME}}

\end{abstract}

\section{Introduction}
\begin{figure}[t!]
    \centering
    \includegraphics[width=.8\linewidth]{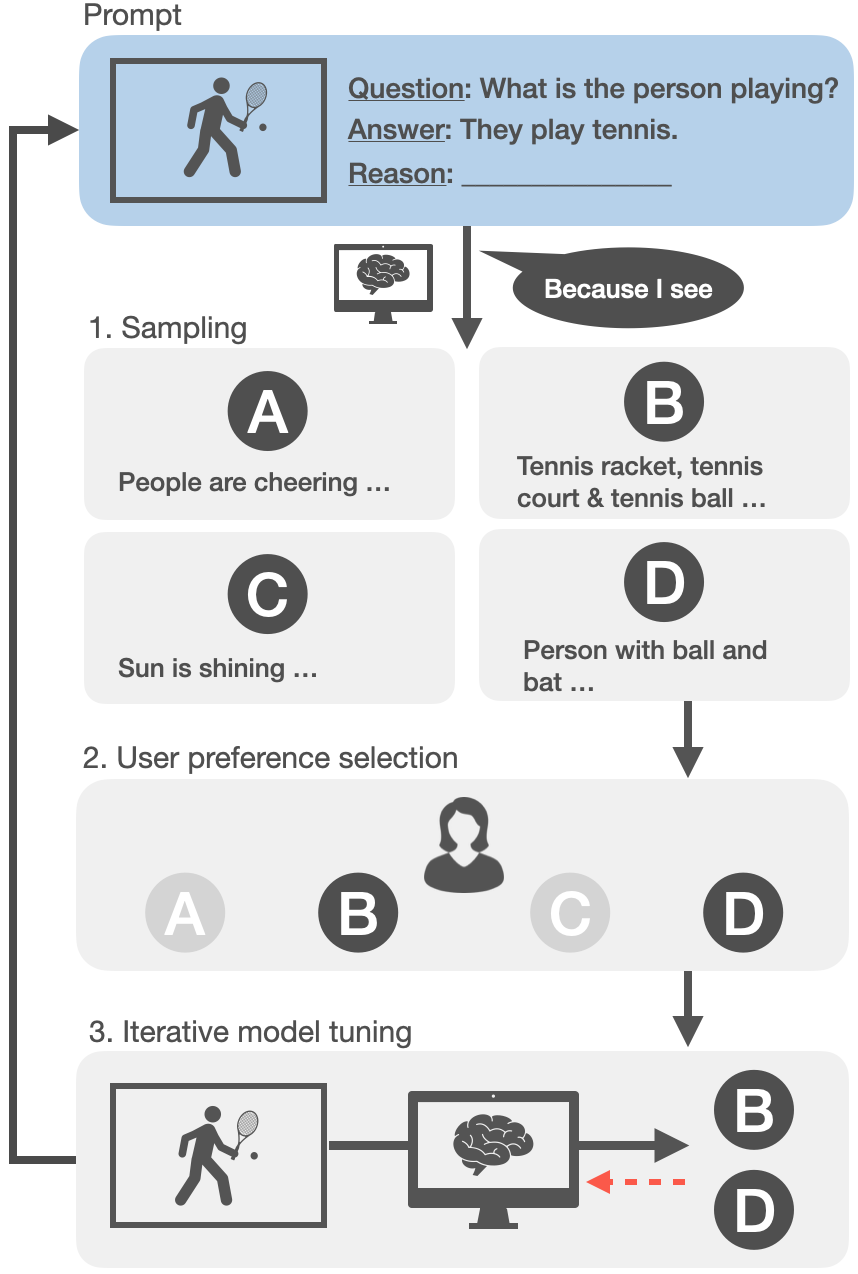}
    \caption{
    \illume~fine-tuning scheme to transfer reasoning capabilities from language models to vision-language models. Based on a VQA input, (1) we sample multiple rationales using VLM, and (2) let an annotator choose \textit{fitting} reasons. 
   (3) The model is fine-tuned---aligned to the human preferences---on all selected rationales where at least one fitting explanation exists. This process is iterated until, for each sample, a fitting reason is generated or no progress can be observed. 
   Note that direct user feedback can be replaced by automatic reward systems. However, this could require prior expensive human labor and is inherently limited.}
    \label{fig:multimodal_selftalk}
    \vskip -0.01in
\end{figure}
Recent vision-language models (VLM) are predominantly built on pre-trained large language models (LM) \cite{tsimpoukelli2021multimodal,eichenberg2021magma,wang2022unifying,li2022blip}. 
However, the behavior of LMs is often not aligned with human intent on a given task \cite{bohmmasani2021opportunities,tamkin2021understanding}. 
Consequently, the bootstrapped VLMs inherit this misalignment.
%
In this work, we propose a human-in-the-loop learning scheme to bridge this gap. 
%
Specifically, we combine pre-existing capabilities inherent to large-scale LMs with interactions on machine generated data. This achieves both transfer of existing commonsense knowledge to the downstream VLM and alignment with human rationalization in a single process. 

This paradigm is called \illume~(\textbf{I}nteractive\textbf{L}y Rationa\textbf{L}izing Vision-Lang\textbf{U}age Mod\textbf{E}ls) as illustrated in Fig.~\ref{fig:multimodal_selftalk}.
During the interactive process, the model's performance improves solely based on machine generated samples from the model itself (see Step 1) selected by human feedback (Step 2), interactively aligning the model to human preferences and gradually carving out rationalization capabilities (Step 3).
Our empirical evaluation demonstrates that \illume~uncovers and reinforces latent capabilities while balancing the benefits of human feedback against the labor-intense generation of ground truth data. 

We target rationales for visual question-answering (VQA) as this is a generic formulation encompassing the majority of vision-language tasks \cite{Manmadhan2020visual}.
Specifically, we contribute: (i) evaluating commonsense reasoning---language and vision---of three recent VLMs, (ii) analyzing the transfer (from language to vision) of their commonsense rationalization capabilities, (iii) introducing a novel iterative tuning paradigm for multimodal architectures solely using human feedback on machine generated data (iv) and demonstrating \illume~on the real world task of providing rationales for immorality in images.

We proceed as follows. First, we briefly discuss related work.
Subsequently, we describe the task and \illume~to transfer reasoning capabilities between modalities. Before concluding and discussing the benefits as well as limitations, we present our experimental evaluation showing indeed LMs can transfer their rationalization capabilities to VLMs using the \illume~paradigm, achieving competitive performance on several open-ended visual reasoning benchmarks and generating satisfactory rationales on a socio-moral task.

\section{Background}

Recently, InstructGPT \cite{ouyang2022training} has demonstrated tuning language models with humans-in-the-loop produces outputs that humans prefer over those of larger, conventionally trained models. Similarly, we use minimal interactive feedback from a human critic on self-generated samples to guide the fine-tuning process. Further, we apply our approach to multimodal applications and facilitate the transfer of capabilities between LMs and VLMs.

Namely, we consider the task of transferring textual reasoning from LMs to vision-language reasoning in VLMs. To elaborate on visual reasoning, recent works have extended upon VQA tasks by considering natural language rationales. 
For instance, \citeauthor{zellers2019from} provide a dataset for visual commonsense reasoning that includes rationale explanations for a VQA task. However, the task is not posed as open-ended generation; instead, both answers and the explanation must be selected from a predefined set of possible options.

In contrast, the Pointing and Justification Explanation model (PJ-X) by \citeauthor{park2018multimodal} generates open-ended textual explanations for VQA and visual heatmaps pointing towards the evidence of an answer. 
Similarly, \citeauthor{Wu2019faithful} proposed the Faithful Multimodal Explanation model (FM), which relies on a pre-existing answering model that is fed a combination of textual and visual representations. 
These architectures are complex and tailored explicitly to perform that one task. In this work, we propose utilizing a pre-trained multimodal VLM instead, offering a more versatile approach and allowing to leverage capabilities of the underlying LM.

\section{Rationalizing Vision-Language Models}
Let us start off by describing the task at hand in more detail before introducing our \illume~approach.
\subsection{Problem Statement}
Recent state-of-the-art vision approaches build models on pre-trained (foundation) LMs \cite{tsimpoukelli2021multimodal,eichenberg2021magma,wang2022unifying,li2022blip,saharia2022photorealistic}.
Here, we aim to align rationalization to human intent while transferring capabilities from LMs to multimodal VLMs. Specifically, we target rationales for VQA since the majority of vision-language tasks can be posed in question form \cite{Manmadhan2020visual}. Consequently, an approach eliciting rationalization for VQA can be considered universal for vision-language in general. Using pre-existing commonsense knowledge of the LM, we only require human feedback on machine generated data. This eliminates the need for the more complex and error-prone task of human ground truth labeling. The majority of current VLM architectures adhere to the same fundamental principles. Two encoders for vision and language project images and text into a joint embedding space. Subsequently, a transformer-based decoder performs autoregressive, open-ended text generation on the encoded multimodal inputs. Often the architecture is based on a pre-trained language encoder-decoder turned into a multimodal model through slight adjustments to the architecture and additional pre-training. We consider the task of transferring rationalization capabilities inherent to the underlying LM to the corresponding VLM. Therefore, we make efficient adjustments to the decoder in order to elicit the desired behavior. 

In this context, we do not treat reasoning as a multiple-choice answer task \cite{kafle2016AnswerType} but as open-ended text generation \cite{tsimpoukelli2021multimodal}. We consider VQA tuples $(i, q, a)$ consisting of an image $i$ and a respective pair of text sequences for the question $q$ and answer $a$. We employ the model to perform a function $f(i,q,a) = e$ that elaborates on the visual question answering and provides a textual explanation $e$.

An explanation refers to an explicitly generated textual sequence $e$ and does not target the interpretability of the model's output. In line with previous research, we use the terms \textit{reasoning} and \textit{explanations} to describe the generation of \textit{rationales} for VQA and use these terms interchangeably.

\subsection{Self-talk Prompting}
Our proposed approach is closely related to the self-talk \cite{shwartz2020unsupervised} prompting paradigm. Instead of interactive transfer of capabilities between modalities, the self-talk approach focuses on improving reasoning via self-generated clarifications. However, we assume that LMs achieving a solid performance on this task are predestined for multimodal (vision-language) reasoning via VLMs.
Therefore, we first establish a baseline for commonsense reasoning in natural language using the self-talk approach to evaluate and, in turn, select fitting LM candidates. 

More precisely, self-talk aims to elicit world knowledge encoded in the model through multi-step prompting. The model is guided towards generating explicit clarification context for the original question that results in more faithful answers. Both clarification and context are prompted to the model to predict the final answer (further details in App.~\ref{appendix:sec:selftalkprompting}).

%
\subsection{\illume: Tuning through Machine Interaction}
\label{sec:mm_reasoning}
For vision-language rationalization, we introduce \illume, a tuning framework that leverages a model's capabilities in one modality and enables transferring these skills to multimodal applications with the human in the loop. 
To that extent, we apply iterative sampling, human feedback, and fine-tuning, as shown in Fig.~\ref{fig:multimodal_selftalk}.
In short, at each iteration, we sample explanations from the training data using the tuned model of the previous iteration. Minimal human feedback is provided to the model through marking \textit{fitting} explanations.


\textbf{Sampling.}
The first step of \illume~is sampling rationale explanations given an input (image-question-answer) prompt. Expressive sampling techniques for LMs have been a long-standing point of discussion in the scientific community. On the one hand, just choosing the most probable token at each position in the sequence may lead to dull outputs. On the other hand, the tail of the distribution of token probabilities might still hold a significant portion of the total probability mass. This makes it inadvertently likely to predict completely unrelated tokens. The most prominent approaches to combat these issues are temperature sampling, \textit{top-k }sampling, and \textit{top-p} aka nucleus sampling. 

Throughout this paper, we rely on the following sampling approach, which combines \textit{top-k} and temperature sampling. First, we apply \textit{top-k} sampling to limit the generated sequence to the most probable tokens. On the filtered tokens, we apply temperature sampling as follows. Consider the logit $l_i$ of the output probability $p_i$ assigned to a token $i$. Temperature sampling scales the logits by $T$ before applying softmax and samples from the resulting distribution:

$$\hat{l_i} = \text{softmax}_I\Big(\frac{l_i}{T}\Big) = \frac{e^{l_i / T}}{\sum_j e^{l_j / T} } \ .$$

Low temperatures push the models toward selecting the most probable tokens, whereas higher temperatures lead to low probability tokens being chosen more often.
Subsequently, we keep $k$ fixed and generate multiple outputs at different temperatures $T \in (0, 1)$ to receive a diverse yet syntactically and semantically more sound set of samples.

Additionally, we aid the sampling process through prior prompt engineering. Initially, we test multiple suitable explanation prompts for each combination of model and dataset. An explanation prompt is the sequence of tokens appended to the image, question, and answer to elicit textual explanations. We evaluate multiple sound options and identify the best scoring prompt(s), which we then use in later sampling. The diversity of samples can be increased even further by repeating the process with multiple explanation prompts. Nonetheless, this comes at the cost of substantially increased computing requirements, and our results indicate that using only the best prompt is sufficient in most cases.

\textbf{Human Feedback.}
A significant portion of generated explanations is likely to be of poor quality, especially in the first iterations. Wherefore, we refer to the unfiltered set of samples as \textit{jabber}. Subsequently, we identify and reinforce those portions of the generated jabber conforming to human intent.
Following sampling, in the second step, a critic labels each explanation as either \textit{fitting} for the image-question-answer pair or \textit{not fitting}. Thus, attenuating the generation of jabber towards on-point explanations. This 
makes \illume~closely related to explanatory interactive machine learning (XIL), in which the human user provides feedback to the training process by interacting with the model's explanations \mbox{\cite{friedrich2022typology}}.

It is noteworthy that at this stage, the human feedback can be simulated by comparing the generated candidates to existing human-generated ground truth explanations using task-specific metrics. For instance, in our experiments, we leverage the ROUGE-L score \cite{lin2004Rouge} to benchmark our approach, i.e., for each explanation candidate, we calculate the sample-wise score between the generated hypotheses and ground truth reference(s). However, this requires prior, labor-intensive human labeling and is 
limited by well-known shortcomings of these approaches. We discuss this further in the empirical evaluation and limitation sections. 

We highlight that this setup is only used to evaluate \illume~itself on well-established benchmarking datasets that allow for empirical comparisons against other approach. Naturally, for the majority of visual-language tasks, ground truth explanations do not exist, and these are the use cases we envision \illume~to be applied on. In addition to the benchmark results, we demonstrate the application of \illume~on a socio-moral task without ground truth data using real human feedback instead (cf. Sec.~\ref{sec:illume_smid}).

\newpage
\textbf{Continual Learning.}
The final step of an \illume~iteration is fine-tuning the VLM based on the selected self-generated samples. As a parameter-efficient approach towards fine-tuning a large neural network, we use bottleneck adapters \cite{houlsby2019Parameter} which are one of the most prominent tuning approaches for continual learning on large-scale transformers. More precisely, we optimize the parameters $\theta$ of small adapter layers inserted at each attention and fully connected module of the decoder instead of tuning the complete model's weights. 
We train the VQA and explanation generation task simultaneously, with the training loss 
\begin{equation}
    \mathcal{L}(X, \theta) = \mathcal{L}_{vqa}(X^{A}, X^{E}, \theta) + b \cdot \mathcal{L}_{exp}(X^{E}, \theta)
    \label{eq:loss}
\end{equation}
being the sum of the language modeling loss for the next token prediction of the answer $\mathcal{L}_{vqa}$ and explanation $\mathcal{L}_{exp}$. $X \supseteq X^{A} \cup X^{E}$ (where $X^{A} \cap X^{E} = \emptyset$) is the training set and $\theta$ the set of optimized parameters of the VLM. The training set $X^{E}_i$ is increased before each feedback iteration $i$. These samples are generated by the VLM's parameters $\theta_{i-1}$ and subsequently filtered by human users or a pre-defined reward function and threshold. 

We observed that adding additional training data from the original VQA task makes the tuning process more robust and also leads to better explanations. Therefore, we add VQA samples without explanation ($X^{A}$) to the training data. In total, the VQA task consists of the VQA pairs of the self-generated training data $X^{E}$ as well as a randomly drawn subset $X^{A}$ of $X \setminus X^{E}$.
We scale the VQA and explanation loss to balance out the disproportional number of samples with 
$b=\frac{n(X^{A})}{n(X^{E})}$, where $n(X)$ is denoted as the number of elements in set $X$. The iterative nature of \illume~makes the tuning setup related to continual learning. We also aim to incorporate additional information into the model with each iteration without losing previous capabilities. Reintroducing a task related to this previous knowledge of the model---in this case VQA---is a standard technique in continual learning. We discuss this further in App.~\ref{appendix:sec:continual_learning}.

The language modeling loss reflects the assumption that the probability $P$ of every token $t_i$ in a sequence can be expressed as the conditional probability of that token given all previous ones:

$$P(t_1, t_2, ..., t_n) = \prod\nolimits_{i=1}^n P(t_i | t_{<i})\ .$$

In an autoregressive neural network, the probability distribution of a token at time step $i$ can be expressed as softmax over all token logits. The loss for predicting this token is the cross entropy between these softmax logits and a one-hot encoding of the target token. Further details, if needed, can be found in \cite{radford2018improving}.

\section{Experimental Results}
Here, our intention is to investigate the transfer of reasoning from natural language to multimodal VQA across three VLMs with distinctive architectural differences. 
Before evaluating the introduced \illume~approach, we compare the rationalization capabilities of the underlying LMs in natural language using self-talk prompting and subsequently establish a correlation with multimodal VQA reasoning.

\subsection{Experimental Protocol}\label{sec:exp_setup}
Let us first clarify the details of our experimental protocol.

\textbf{Models.}
We consider three recent VLMs, which differ mainly in the choice of LM on which to build the multimodal model. 1) MAGMA \cite{eichenberg2021magma}, whose LM-foundation is a large GPT model, 2) BLIP \cite{li2022blip}, which uses BERT (a less powerful initialization), and 3) OFA \cite{wang2022unifying}, which is trained from scratch.

In Sec.~\ref{sec:exp_nlp_selftalk} we investigate the underlying language models of each VLM. For MAGMA, we consider luminous-base, which itself is based on the GPT-J architecture.
Further, we evaluate the base version of BERT as the underlying language model of BLIP. Since OFA is trained from scratch, no baseline language model exists to consider. Instead, we evaluated the large general pre-trained OFA checkpoint, using it only with natural language sequences. 

Based on the experiments in Sec.~\ref{sec:exp_nlp_selftalk} and \ref{sec:exp_zero_shot_transfer}, MAGMA has proven as most suitable for \illume. Hence, we continue the subsequent evaluation solely on MAGMA. Subsequently, we refer to the zero-shot model as MAGMA$_\text{base}$ to distinguish it from fine-tuned variants. 

\textbf{Datasets \& Benchmarks.}
We use six diverse commonsense reasoning benchmarks to evaluate self-talk in natural language. These datasets are CSQA~\cite{talmor2019commonsenseQA}, COPA~\cite{gordon2012semeval,roemmele2011choice}, McTaco~\cite{zhou2019going}, PIQA~\cite{bisk2020piqa}, Social-IQA~\cite{sap2017social} and WinoGrande~\cite{sakaguchi2020winogrande} which cover a wide range of reasoning tasks ranging from basic real-world concepts to physical and social interactions as well as temporal commonsense. All datasets provide multiple-choice answers, with a model's performance being measured as its accuracy in choosing the correct one. 

For the visual reasoning task we consider three datasets, namely VQA-X, 
ACT-X~\cite{park2018multimodal}, and \mbox{CLEVR-X}~\cite{DBLP:conf/icml/SalewskiKLA20}.
Contrary to the reasoning benchmarks in natural language, we treat multimodal reasoning as open-ended text generation without providing multiple-choice alternatives.
Further details on the composition of these datasets can be found in App.~\ref{appendix:sec:datasets}.
Additionally, we provide benchmark results of MAGMA on these datasets compared to the current state-of-the-art models in App.~\ref{appendix:sec:gtexp}. 

Further, we observed that prompt engineering of the explanation prompt can significantly effect explanation quality. For comparisons between models, we evaluate each model with the same set of potential explanation prompts and report the scores for the best-performing one. Since ACT-X is not originally framed as a VQA task we engineered a similar set of questions for the dataset and evaluate every combination of questions and explanations for each model.
In addition, similarly to \cite{park2018multimodal}, we observed that the quality of explanations depends on the answer given in the context prompt. Therefore, we used the ground truth answer instead of the model-generated one for all experiments to allow a fair comparison between models.

\textbf{\illume: Sampling.}
We performed sampling using the VLM on the training data to generate five explanations, each at five different temperatures. We set $k=5025$ to be equal to 10\% of the vocabulary size and select temperatures \mbox{$T \in \{0.01, 0.1, 0.3, 0.6, 0.9\}$}. In total, this yields up to 25 different explanations. For \mbox{ACT-X}, we additionally sampled with five questions per image resulting in 125 samples total. 
The explanations generated in this manner were very diverse, with most of the resulting jabber being distinctly unsuitable for further fine-tuning. Nevertheless, this approach is intended to generate a large variety of samples to increase the likelihood of generating \textit{fitting} ones. However, this also requires the generated explanations to be filtered rigorously.

\begin{table*}[t]
    \small
    \centering
    \def\arraystretch{1}\tabcolsep=8.pt
    
    \begin{tabular}{l  r r r r r r}
    
    \textbf{Model} & \textbf{CSQA$\uparrow$} & \textbf{COPA$\uparrow$} & \textbf{MC-TACO$\uparrow$} & \textbf{PIQA$\uparrow$} & \textbf{Social-IQA$\uparrow$} & \textbf{WinoGrande$\uparrow$} \\ \hline
     \textit{Chance} & \textit{20.0} & \textit{50.0} & \textit{18.9} & \textit{50.0} & \textit{33.3} & \textit{50.0} \\ \hline
     GPT-J-6B   & $\mathbf{51.7}\bullet$ & $\mathbf{74.0}\bullet$ & $\mathbf{64.8}\bullet$    & $\mathbf{71.2}\circ$ & $\mathbf{46.3}\circ$       & $\textbf{59.9}\circ$ \\ \hline
     BERT (BLIB)     & 21.5 & 64.0 & 39.0    & 48.1 & 32.8     & 49.1 \\
     OFA-LM (OFA)    & 17.8 & 53.0 & 43.6    & 51.9 & 34.0     & 50.7 \\  
     Luminous-base (MAGMA) & $\mathbf{45.5}\circ$ & $\mathbf{72.0}\circ$ & $\mathbf{62.3\circ}$ & $\mathbf{77.4}\bullet$ & $\mathbf{47.2}\bullet$ & $\mathbf{62.1}\bullet$\\
        
    \end{tabular}
    \caption{LM's self-talk performances. Question answering accuracy (\%) of models are reported on the dev. sets of 6 commonsense multiple-choice QA tasks. Higher scores are better. All models use self-talk as a knowledge source. 
    The \textit{chance} row represents the expected accuracy achieved by selecting a multiple-choice answer randomly. The best (``$\bullet$'') and runner-up (``$\circ$'') results are highlighted \textbf{bold}. }
    \label{tab:nlp_reasoning}
    \vskip -0.1in
\end{table*}
\begin{table}[t]
    \centering
    \def\arraystretch{1}\tabcolsep=3.pt
    \small
    \begin{tabular}{l c c c | c c c | c c c}
    & \multicolumn{3}{c}{\textbf{VQA-X$\uparrow$}}   &  \multicolumn{3}{c}{\textbf{ACT-X$\uparrow$}}   &  \multicolumn{3}{c}{\textbf{CLEVR-X\footnotemark$\uparrow$}} \\ 
    \textbf{Model}     & \textbf{B-4} & \textbf{R-L}  &   \textbf{C}          & \textbf{B-4} & \textbf{R-L}     &   \textbf{C}       & \textbf{B-4} & \textbf{R-L}    &   \textbf{C}  \\ \hline
     OFA        & \phantom{1}0.3  & \phantom{1}9.1  & \phantom{1}8.5  & 0.3  & 11.8  & \phantom{1}7.2  & \phantom{1}0.0   & \phantom{1}2.9   & \phantom{1}0.5   \\
     BLIB       & \phantom{1}0.0  & \phantom{1}5.5  & \phantom{1}6.9  & 0.0  & \phantom{1}6.9  & \phantom{1}4.8  & \phantom{1}0.0   & \phantom{1}3.2   & \phantom{1}0.5   \\
    MAGMA     & $\textbf{\phantom{1}9.2}$  & $\textbf{32.5}$ & $\textbf{31.1}$  & $\textbf{3.3}$  & $\textbf{22.4}$  & $\textbf{17.1}$  & $\textbf{23.1}$  & $\textbf{49.4}$  & $\textbf{19.7}$

    \end{tabular}
    \caption{Zero-shot reasoning performance. Results are reported on the respective validation datasets. Scores refer to Bleu-4, Rouge-L \& CIDEr where higher scores are better and best results are \textbf{bold}. Explanations are generated conditioned on the ground truth answer. Scores are reported for the best performing prompt for each combination of model and dataset. Please note that total scores are not directly comparable between datasets as they are heavily influenced by the number of provided references as well as their vocabulary size and sequence length \protect\cite{DBLP:conf/icml/SalewskiKLA20}. 
    Both of these factors vary significantly between the datasets making meaningful, direct comparisons impossible.}
    \label{tab:zeroshot_multimodal_reasoning}
    \vskip -0.1in
\end{table}

\textbf{\illume: Feedback.} On the two benchmark datasets we simulated human feedback as follows. We calculated the sample-wise ROUGE-L score between the generated hypotheses and human-annotated ground truth (GT) reference(s). As the quality of an explanation is subjective to some extent (cf.~Sec.~\ref{sec:exp_zero_shot_transfer} and \ref{sec:limitations}) there exists no single \textit{correct} explanation. Therefore, we empirically chose (cf.~App.~\ref{appendix:sec:filter_threshold}) a threshold of \mbox{$\textrm{ROUGE-L} \geq 0.7$} to be a good approximation of \textit{fitting} explanations. We observed that explanations below that threshold are often nonsensical in that they are semantically or syntactically incorrect, incomplete, or simply too different from the ground truth to be a \textit{fitting} explanation.

Within the inherent limits of an automated metric, we deem this to be a reasonable trade-off between addressing differences in wording and filtering out ill-formatted text sequences, thereby turning jabber into sound explanations.

\textbf{\illume: Tuning.} We tuned the VLM (MAGMA) by optimizing the adapter weights (see \cite{houlsby2019Parameter}) contained in the LM transformer of the network, keeping the image prefix module frozen. For all experiments, we used the AdamW optimizer and a batch size of 256. The training was distributed over 8 A100 GPUs resulting in a per GPU batch size of 32. 
Regarding Eq.~\ref{eq:loss}, we added roughly ten times more samples without explanation $X^{A}$ than $X^{E}$ to regularize optimization.
Any additional hyper-parameter optimization was performed on the dedicated validation splits, with the test splits being evaluated only for reporting final scores. We provide further insights on the hyperparameter selection of tuning with self-generated examples in App.~\ref{appendix:sec:self-training}.

\textbf{Evaluation Metrics.}
\footnotetext{Using a 10k random subset of the validation set.}
We use automated natural language generation (NLG) metrics for text generation to assess a model's performance on explanation generation. For references, we rely on the provided ground truth explanations in the datasets. This approach is considered best practice in this area of research.
However, these metrics have well-known limitations that should be considered when relying on them for evaluation \cite{sai2022Survey}. 
First, n-gram based metrics are generally incapable of bridging the semantic gap. Therefore, generated sequences that convey the same meaning but are phrased differently will receive low scores. Additionally, \textit{fitting} explanations are not unique, and a model might generate a suitable explanation that is not included in the references and will thus be discarded. Explanations are subjective to some extent which may be ill-represented in ground truth labels. Case studies comparing human preferences to automated ratings concluded that the scores of all such metrics are not significantly correlated with human rating \cite{novikova2017association}. This observation is especially true for distinguishing between mediocre and good-quality generated sequences. Therefore, comparisons of benchmark scores between multiple decently performing models notably lack significance. However, the authors concluded that these metrics can still provide valuable insight in identifying cases of poor performance and the initial development of a system. Therefore, we deem these metrics good enough to provide empiric evidence of the validity of our approach. 
Subsequently, we report BLEU-4, ROUGE-L, and CIDEr scores for all conducted experiments, which provide a variety of profound insights. Further scores are provided in  App.~\ref{appendix:sec:extmetrics}.

\subsection{Self-talk Prompting}
\label{sec:exp_nlp_selftalk}
We start by analyzing the underlying LMs of BLIP, MAGMA, and OFA\footnote{We note that the official OFA implementation does not support nucleus sampling as proposed for self-talk prompting. Instead, we used the implemented beam-search with the beam width matching the number of samples generated through nucleus sampling.} on the datasets mentioned above.

Tab.~\ref{tab:nlp_reasoning} shows the reasoning performance of the corresponding LMs for each of the considered multimodal architectures. Additionally, we included a popular and publicly available GPT model for reference. 
The GPT-based models, GPT-J and Luminous, outperform weaker pre-trained language models such as BERT and purely multimodal models such as OFA across all tasks. For most datasets BERT and OFA barely---if at all---beat randomly, selecting an answer by chance. These results illustrate the complexity of commonsense reasoning tasks, which are far from trivial. Instead, these problems require fundamental world knowledge and language understanding that are usually only achievable by leveraging large pre-trained models.


\subsection{Zero-Shot Visual Reasoning}
\label{sec:exp_zero_shot_transfer}
In addition to the commonsense abilities of VLMs' underlying LMs, the VLMs' zero-shot performances indicate the portion of reasonable rationales that can be expected among the generated jabber. Therefore, we require a pre-trained model to perform decently on these benchmarks in order to produce a sufficient number of \textit{fitting} explanations that may be used for further fine-tuning.
To this end, we now benchmark the initial, i.e., without additional fine-tuning, multimodal rationalization capabilities of the discussed VLMs.

Tab.~\ref{tab:zeroshot_multimodal_reasoning} depicts the zero-shot reasoning performance of all models.
It is apparent that those VLMs whose LMs perform weak on NLP reasoning also yield low-quality multimodal explanations. However, MAGMA, which is based on a GPT variant with good language reasoning capabilities, can generate decent zero-shot, multimodal explanations 
without any training for that particular task. An example highlighting these differences is depicted in Fig.~\ref{fig:vqax_model_comp}. As is apparent for these inputs, OFA and BLIP tend to overfit on the VQA task, resulting in these models only repeating the answer if prompted for further outputs. On the VQA-X validation set, when prompted for a rationale, OFA and BLIP repeat the answer in 63\% and 89\%  of all samples, respectively. Therfore, we use MAGMA for all subsequent experiments.


\begin{figure}
    \centering
    \includegraphics[width=.9\linewidth]{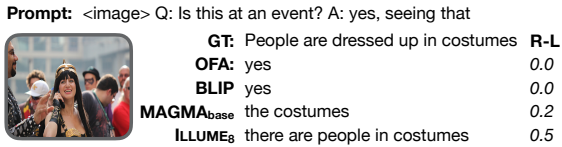}
    \caption{Exemplary comparison of explanations generated on the VQA-X validation set by different models. VQA image, question, answer, and a generated explanation of each model with the ROUGE-L score wrt. ground truth. Explanations for MAGMA$_{\textrm{base}}$, OFA \& BLIP are generated zero-shot. (Best viewed in color) \label{fig:vqax_model_comp}}
    \vskip -0.1in
\end{figure}

\subsection{\illume~-- Benchmark}\label{sec:illume_smid}
Affirmed by the zero-shot capabilities, we applied our \illume~paradigm to the VQA-X and ACT-X datasets. The application of logical reasoning in the form of the \mbox{CLEVR-X} dataset remains challenging, which we discuss in further detail in the limitations (cf.~Sec.~\ref{sec:lim_logic}).

\begin{table}[t]
    \centering
    \small
    \def\arraystretch{1.2}\tabcolsep=1.4pt
    \begin{tabular}{c | l c @{\hskip 7pt} c @{\hskip 7pt} c @{\hskip 4pt} c}
    \multicolumn{6}{c}{\textbf{VQA-X}} \vspace{0.25em}\\
     &\textbf{Iteration} & \textbf{B-4$\uparrow$ $\Delta$} & \textbf{R-L$\uparrow$ $\Delta$} & \textbf{C$\uparrow$ $\Delta$} & \textbf{RV (\%)}  \\
     \hline
     \parbox[t]{3mm}{\multirow{6}{*}{\rotatebox[origin=c]{90}{validation}}}
     & \textit{MAGMA$_{\text{base}}$} &  \phantom{1}9.16 \phantom{$-$0.0} & 32.45 \phantom{$-$0.0} & 31.08 \phantom{$-$0.0} & \phantom{11}0.0  \\
     & \textit{It 1} & 14.06 \textbf{$+$0.2} & 39.52 \textbf{$+$0.3}  & 44.57 \textbf{$+$3.4} & \phantom{11}4.1\\
     & \textit{It 3} & 17.42 $-$1.2  & 42.49 \textbf{$+$0.1}  & 52.91 $-$1.0 & \phantom{11}8.3\\
     & \textit{It 5} & 19.35 $-$0.8  & 43.67 $-$0.2 & 59.51 $-$0.7 & \phantom{1}10.1\\
     & \textit{It 7} & 20.13 $-$0.5  & 44.55 \textbf{$+$1.2}  & 62.85 \textbf{$+$1.2} & \phantom{1}11.5\\
     & \textit{It 8} & 20.86 \textbf{$+$0.7} & 44.75 \textbf{$+$1.2}  & 65.20 \textbf{$+$1.8} & \phantom{1}12.0\\
     \hline
     \parbox[t]{2mm}{\multirow{2}{*}{\rotatebox[origin=c]{90}{test}}} & \textit{It 8} & 19.01 \textbf{$+$0.2} & 44.24 \textbf{$+$0.7} & 60.18 \textbf{$+$2.7} & \phantom{1}12.0\\  \cline{2-6}
     & \textit{MAGMA$_{\text{full}}$} & 21.94\phantom{ $-$0.0}  & 46.76\phantom{ $-$0.0}  & 73.79 \phantom{ $-$0.0} & 100.0 \vspace{0.33cm} \\
    %
    
    \multicolumn{6}{c}{\textbf{ACT-X}} \vspace{0.25em}\\
   &\textbf{Iteration} & \textbf{B-4$\uparrow$ $\Delta$} & \textbf{R-L$\uparrow$ $\Delta$} & \textbf{C$\uparrow$ $\Delta$} & \textbf{RV (\%)}  \\ \hline
    \parbox[t]{3mm}{\multirow{6}{*}{\rotatebox[origin=c]{90}{validation}}}
     & \textit{MAGMA$_{\text{base}}$} &  \phantom{1}3.30 \phantom{$-$0.0} & 22.44 \phantom{$-$0.0} & 17.08 \phantom{$-$00.0} &  \phantom{11}0.0  \\
     & \textit{It 1} & \phantom{1}3.63 $-$6.1 & 27.17 $-$7.2 & 24.81 $-$26.3 & \phantom{11}0.7\\
     & \textit{It 3} & \phantom{1}7.58 $-$5.6  & 33.79 $-$4.8  & 46.98 $-$28.7 & \phantom{11}8.4\\
     & \textit{It 5} & \phantom{1}9.54 $-$4.7  & 35.55 $-$4.1  & 58.39 $-$24.0 & \phantom{1}13.7\\
     & \textit{It 7} & 10.66 $-$3.2  & 36.22 $-$2.9  & 62.84 $-$19.7 & \phantom{1}16.9\\
     & \textit{It 9} & 10.65 $-$2.0  & 36.37 $-$1.1 & 65.02 $-$\phantom{1}7.2 & \phantom{1}18.9\\
     \hline
     \parbox[t]{2mm}{\multirow{2}{*}{\rotatebox[origin=c]{90}{test}}}&\textit{It 9} & 10.79 $-$2.4  & 36.13 $-$1.8  & 64.70 $-$13.2 & \phantom{1}18.9 \\  \cline{2-6}
     &\textit{MAGMA$_{\text{full}}$} & 15.36 \phantom{$-$1.1}  & 40.34 \phantom{$-$1.1}  & 92.96 \phantom{$-$11.1} & 100.0
    \end{tabular}
    \caption{Iterative process of \illume~on VQA-X (top) and ACT-X (bottom) until scores plateau on the validation set. 
    $\Delta$ values next to the scores indicate the difference between training on self-generated samples vs. the same amount of GT samples, with positive scores indicating that \illume~outperforms training on GT (\textbf{bold}) and vice versa.
    MAGMA$_{base}$ refers to zero-shot (\textit{It 0}) performance and MAGMA$_{full}$ refers to the model tuned on the entirety of the GT training set, which are 29459 and 12607 for VQA-X and ACT-X, respectively.
    Additionally, RV displays the relative value wrt. total amount of samples in original training set. 
    The bottom rows show scores on the test set. 
    Bleu-4, Rouge-L \& CIDEr scores are shown (higher is better).   \label{tab:exp_selftaught}}
    \vskip -0.1in
\end{table}
Tab.~\ref{tab:exp_selftaught} shows the progress of \illume~on VQA-X and \mbox{ACT-X}. 
Overall, \illume~generalizes well to unseen data. At the initial iterations, especially on ACT-X, tuning for a single epoch on a small training set significantly increases the number of \textit{fitting} explanations the model generates on new data. We can observe that explanation generation improvements are closely correlated to the number of new samples added to the training data. The number of samples and the NLG scores improve rapidly in the beginning and slowly converge in later iterations. 
Additionally, we can observe \illume~to be more robust against overfitting than tuning with ground truth data. The latter approach suffers a significant drop in scores achieved on the validation set at a stage in the procedure at which the \illume~variant still improves, cf.~CIDEr score in iteration 7 through 9 on ACT-X.
For both experiments, we make the empirical observation that the best scores are achieved once the ratio of new samples drops below 5\%, e.g., the number of samples for VQA-X from iteration 7 to 8 only increases from 3385 to 3541, equaling 4.6\%. Therefore, this threshold might be a vital indicator for performance saturation in datasets without ground truth reference. 

\begin{figure}[t]
    \centering
    \includegraphics[width=.9\linewidth]{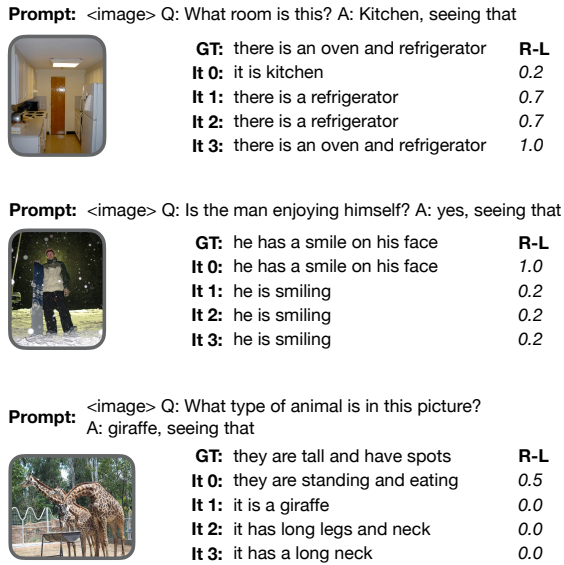}
    \caption{Generated explanations on the VQA-X training set. Image, question, answer, and a ground truth explanation are shown. On the bottom, we depict the best generated explanation and ROUGE-L score wrt. GT at every iteration. (Best viewed in color)\label{fig:vqax_exampleAll}}
    \vskip -0.1in
\end{figure}

More precisely, in the case of VQA-X, the quality of explanations improves for eight iterations until the scores plateau.
The resulting \illume~model even slightly outperforms the model obtained through standard supervised learning on ground truth data. Additionally, \illume~yields a model remaining competitive with MAGMA$_{full}$ while using no ground truth explanations and less data.

In the case of ACT-X, we had to apply slight modifications to address the nature of the dataset. The number of \textit{fitting} explanations generated in a zero-shot fashion is significantly lower than for the other datasets. We addressed this issue by sampling the training set with multiple question prompts and two different explanation prompts. For the initial sampling, this significantly boosts the number of \textit{fitting} explanations. The benefit of using more than one explanation prompt for sampling diminishes with subsequent iterations as the model is conditioned on the prompt used in training. Therefore, we only employed it for the first sampling iteration. 
Nonetheless, the initial number of samples remains comparatively low, making up less than 1\% of the ground truth training set.
Further, while fine-tuning the VLM on the ACT-X ground truth data, we observed that training on only one fixed question might lead to unstable training behavior, especially on smaller subsets of the training set. Therefore, we chose to use five different---albeit similar---question prompts during the training of both the VQA and the explanation task. 
This adjustment makes the \illume~self-generated data more diverse and leads to more robust training.

In summary, our empirical results clearly show that \illume~achieves competitive performance and can incorporate direct human feedback, making it a more effective approach for tuning foundation models than using truth data. Note that this only applies to tasks on which the model displays rudimentary capabilities through language or multimodal pre-training; see results on CLEVR-X in Sec.~\ref{sec:lim_logic}.

\begin{table}[t]
    \centering
    \small
    \begin{tabular}{c | l  @{\hskip 12pt} c @{\hskip 12pt} c}
    \multicolumn{4}{c}{\textbf{SMID}} \vspace{0.25em}\\
    & &  \multicolumn{2}{c}{\textbf{Accuracy (\%)} $\uparrow$} \\
    & \textbf{Iteration} & Rating 1 & Rating 1-2  \\
     \hline
     \parbox[t]{3mm}{\multirow{4}{*}{\rotatebox[origin=c]{90}{Test}}} & \textit{MAGMA$_{\text{base}}$} & \phantom{0}6.06 & 13.64 \\
     & \textit{It 1} & 12.12 & 63.64 \\
     & \textit{It 2} & 39.39 & 66.67 \\
     & \textit{It 2*} & \textbf{42.24} & \textbf{77.27} \\
     
    \end{tabular}
    \caption{Percentage of fitting rationales being generated on the held-out SMID test set over continuous \illume~iterations. It 2* denotes the model tuned from the start using all training samples accumulated over 2 iterations.}
    \label{tab:smid}
    \vskip -0.15in
\end{table}

\subsection{\illume~-- Real World}
Going beyond existing benchmarks, we employed \illume~on a dataset with no ground truth rationales and provided real human feedback instead. We chose to apply the approach on explanations for immoral image content. While classification methods proposed in previous work \cite{schramowski2022can} provide binary decision whether images can be considered immoral, there is a severe lack of rationales in existing methods. For this task we used the Socio-Moral-Image-Database (SMID) \cite{crone2018socio} that contains images rated on their immorality by human users. For \illume~we consider those images with a mean moral score below $2.0$ (out of 5). The images were split into a training set on which we supplied feedback and a test set for evaluation containing 208 and 66 photos, respectively.

Similar to the \illume~prompting paradigm employed for ACT-X, we used 12 different Q/A prompts in the form of `\textit{Q: Is the image content immoral? A: Yes, because...}' To facilitate a more meaningful evaluation, we rated each generated explanation from 1 - 5, with 1 being the best score. Ratings 1 - 4 correspond to the generated explanation being \textit{excellent (1)}, \textit{sufficient/satisfactory (2)}, \textit{weak (but right direction) (3)}, and \textit{poor/unrelated (4)} whereas automatically pre-filtered jabber is scored as 5. Explanations rated as 1 or 2 were labeled as fitting for tuning.

We report the test accuracy for fitting explanations in Tab.~\ref{tab:smid}. Again, Again, the model improves its performance significantly with passing \illume~iterations. Additionally, we observed the model to generate rationales of higher quality if we train the model for one iteration using the entirety of samples accumulated over all iterations. We denote this model version as It2*. This quality difference can be attributed to the fact that the iterative setup does not include all samples the same number of times, leading to overfitting on samples already included in early iterations. Interestingly, this effect is only observed for the human rated quality of rationales and not on the automatically evaluated NLG metrics from previous experiments. 
We further visualize the gradual quality shift on the training and test set in Fig.~\ref{fig:smid_scores}. The base model generates jabber (rating $1$) for the vast majority of samples. However, after tuning for only one iteration the number of fitting explanations exceed the non-fitting ones. After further iterations the model mostly generates fitting explanations on the training data and yields higher quality rationales on the test set. 

\newpage
We make the labeled rationales 
available online to facilitate further research\footnote{
\url{https://huggingface.co/datasets/AIML-TUDA/socio-moral-image-rationales} }.

\begin{figure}
    \centering
    \includegraphics[width=.9\linewidth]{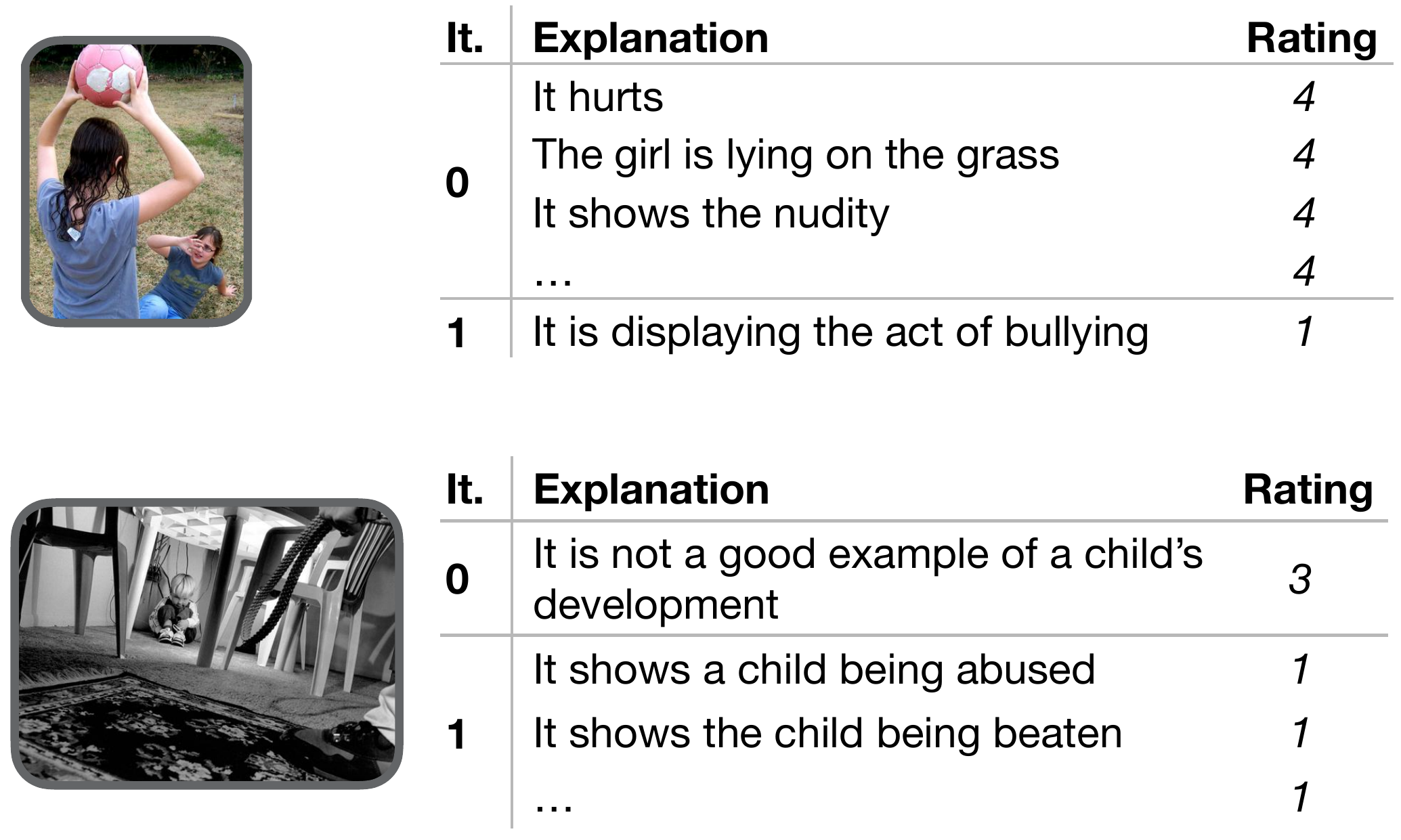}
    \caption{Generated explanations on the SMID training set. Image and generated explanations are shown. For both the base model (iteration 0) and the ILLUME model (iteration 1), we depicted the best explanations based on the human rating. In both cases no explanation generated from the base model is used during training. (Best viewed in color)}
    \label{fig:smid_examples}
    \vskip -0.15in
\end{figure}

\section{Discussion \& Limitations}\label{sec:discussion}\label{sec:limitations}\label{sec:lim_logic}
Before concluding, we discuss the transfer and progressive alignment of VLMs' reasoning capabilities in more detail. 
Furthermore, we touch upon limitations and observed shortcomings of \illume.

\textbf{Progressive Explanation Alignment.}
The NLG metrics used to automatize feedback provide high-level information on the iterative progress of aligning generated explanations to ground truth ones. Nonetheless, an additional qualitative evaluation of the process can provide valuable insights. 

Fig.~\ref{fig:vqax_exampleAll} (top) depicts an example of the VQA-X training set representative for explanation improvements with passing \illume~iterations. Initially, the model is likely not to generate a concise explanation. Instead, it produces text that either resembles a caption of the image or repeats the answer from the prompt. After one training iteration, the model generates a reasonable fitting explanation, and two iterations later, the output is equal to the dataset's ground truth. It is important to note that this improvement is inferred from other (self-generated) training examples as the actual ground truth data is never presented. Overall, the model generalizes well between different samples of commonsense reasoning. 
The same generalization capabilities be can be observed for the morality experiment as shown in Fig.~\ref{fig:smid_examples}. \illume~infers rationales for complex concepts like bullying and child abuse from human feedback on other training samples.

\begin{figure}
     \centering
     \begin{subfigure}[t]{0.15\textwidth}
         \centering
         \includegraphics[height=85pt]{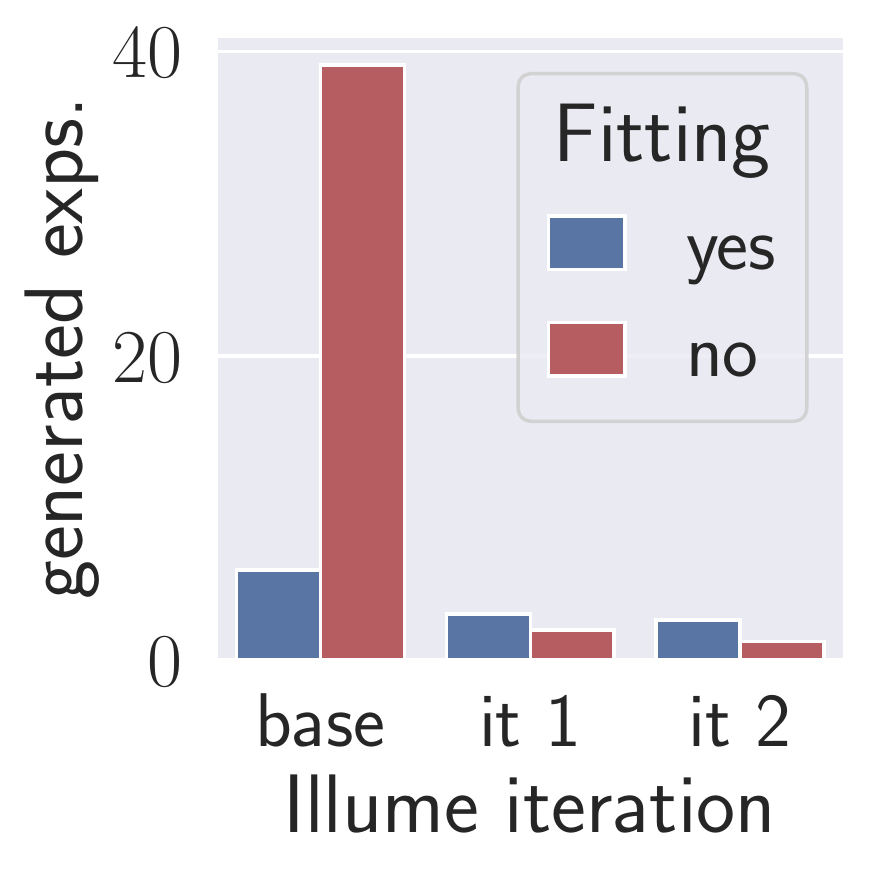}
         \caption{Avg. number of additional explanations per image in the train set}
         \label{fig:smid_scores_train}
     \end{subfigure}
     \hfill
     \begin{subfigure}[t]{0.30\textwidth}
         \centering
         \includegraphics[height=85pt]{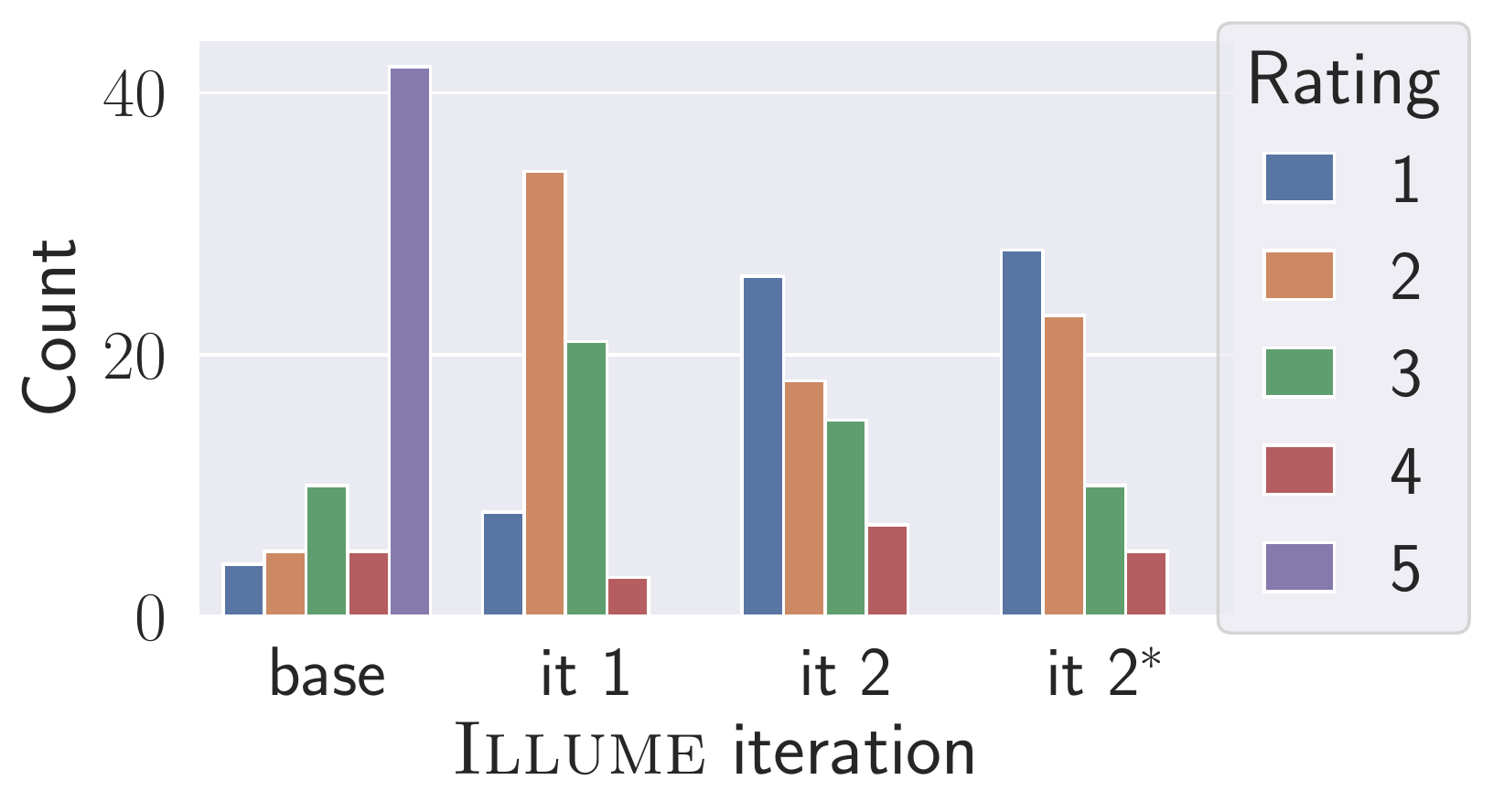}
         \caption{Test set distribution over the samples’ best explanation-rating 
         }
         \label{fig:smid_scores_test}
     \end{subfigure}
        \caption{\illume’s tuning process illustrated on the SMID train (a) and test set (b). \illume~is able to identify fitting explanations among a large number of jabber. \\(Best viewed in color)}
        \label{fig:smid_scores}
        \vskip -0.1in
\end{figure}

\textbf{Limitations of \illume.}
Adapter tuning \cite{houlsby2019Parameter} is an efficient approach to fine-tune large models. However, exploring other optimization approaches can provide a more holistic set of tools with potential use in different scenarios. More recent parameter-efficient finetuning techniques could reduce the computational requirements even further \cite{dettmers2023qlora}. We observed continuous prompt-tuning \cite{lester2021thepower} to be such a promising candidate. Initial experiments on optimizing the embedding of the explanation prompt without adjusting any parameter weights yielded positive results. 

Additionally, prompt-tuning could be one viable solution in tackling the problem of needing to tune a dedicated reasoning model, here a set of adapters, for each dataset. In any regard, a general model for reasoning would be preferable.

Furthermore, we would like to reiterate the issues of automatic NLG metrics. Fig.~\ref{fig:vqax_exampleAll} (middle) provides an example of metrics failing to bridge the semantic gap. The sentences \textit{'he has a smile on his face'} and \textit{'he is smiling'} are scored as substantially dissimilar, although they are semantically identical. Yet another example is shown in Fig.~\ref{fig:vqax_exampleAll} (bottom). Here the generated explanation \textit{'they are standing and eating'} is rated significantly higher than \textit{'it has long legs and neck'}, although the first one provides virtually no valid information on why the animal is a giraffe, whereas the second one identifies two of its most prominent features. 
This further illustrates the limited significance of comparisons between models using automatic NLG metrics. 
Nevertheless, as described, such metrics are a valid indicator to evaluate a method itself. Hence, we benchmarked \illume~on several datasets utilizing ROUGE-L to simulate user feedback and a wide range of scores for evaluation. Yet, the above-discussed examples further motivate \illume's intended use of direct \textit{human} feedback in training and evaluation.

\textbf{Flaws in Logical Reasoning.}
One frequently observed shortcoming of large neural networks is their inability to generalize to logical reasoning. \citet{zhang2022paradox} recently demonstrated that BERT does not learn logical reasoning but instead captures statistical features in the training data. Therefore, the model remains unable to generalize to other distributions of the exact same problem. In the multimodal domain, DALL-E 2 \cite{ramesh2022hierarchical} fails to construct logical relations between objects faithfully. 

We also observed \illume~to yield no satisfying results on CLEVR-X. Details i can be found in App.~\ref{appendix:sec:logical}.
Summarized, we attribute this behavior to the same observations made by \citeauthor{zhang2022paradox} in that current LMs appear incapable of inferring logical reasoning from a few training examples. Therefore, VLMs bootstrapped from LMs struggle to transfer logical reasoning capabilities without major extensions.
Instead, we argue that the approach of training and evaluating logical reasoning as a pure text generation task may be inherently flawed. Instead, logic-based methods \cite{shindo2021neuro-symbolic} using differentiable forward-chaining using first-order logic could yield more coherent explanations. 

\section{Conclusion}
We proposed \illume, a human-in-the-loop rationalization approach for multimodal transformers. As our experiments demonstrate, \illume~enables the transfer of commonsense reasons from LMs to downstream VLMs. 
In particular, the \illume~approach remains competitive with approaches that train on substantially larger datasets that were previously labeled without the human out of the loop.
Further, it paves the way toward lowering the workload on annotators and enables aligning the model to users' rationales through interactive feedback in the training loop. 

Our paper provides several avenues for future work, including the ones discussed above. Further, an increasing number of powerful LMs has been released publicly since the writing of this work. These models create further opportunities for investigating the capability transfer to other modalities.
Probably the most important one is future work on transformers' logical and commonsense reasoning capabilities.


\section*{Acknowledgments}
 We gratefully acknowledge support by 
 the German Center for Artificial Intelligence (DFKI) project “SAINT” and 
 the Federal Ministry of Education and Research (BMBF) under Grant No. 01IS22091. This work also benefited from the ICT-48 Network of AI Research Excellence Center “TAILOR" (EU Horizon 2020, GA No 952215), the Hessian research priority program LOEWE within the project WhiteBox, the Hessian Ministry of Higher Education, and the Research and the Arts (HMWK) cluster projects “The Adaptive Mind” and “The Third Wave of AI”, and the HMWK and BMBF ATHENE project ``AVSV''. Further, we thank Felix Friedrich for his valuable feedback.
\end{alphafootnotes}

\bibliographystyle{named}
\bibliography{bibliography}

\clearpage
\begin{figure*}[tbh!]
    \centering
    \includegraphics[width=0.8\linewidth]{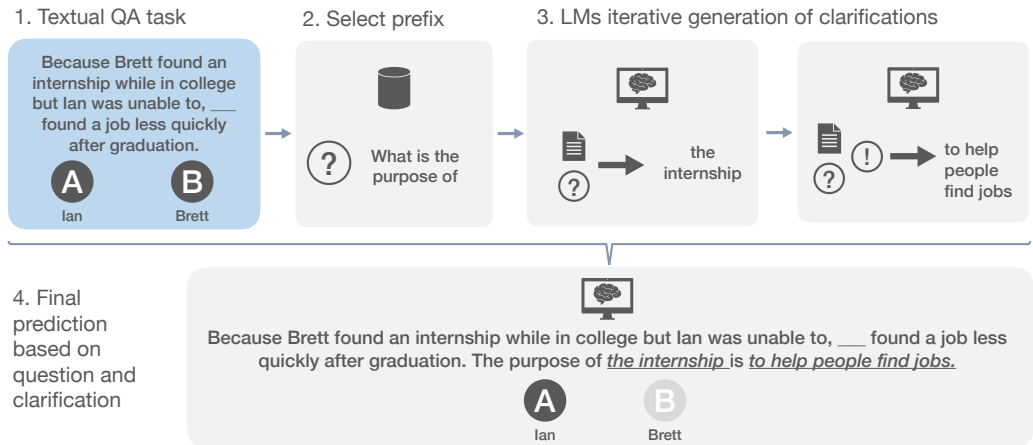}
    \caption{
    Pipeline for generating clarifications through self-talk prompting \protect\cite{shwartz2020unsupervised}. (1) Each of the original questions or promises (2) is concatenated with multiple question prompts specifically engineered for each dataset. For simplicity's sake figure only shows one prompt. 
    (3) For each task context and question prompt the model generates completions for the posed question. The original, prompt and the completed clarification question are concatenated and fed to the model again. From that prompt the model completes the clarification answer (4) which is in turn used as context for solving the task.
    }
    \label{fig:selftalk}
\end{figure*}
\appendix

\section{Natural Language Self-Talk Prompting}
\label{appendix:sec:selftalkprompting}
Self-talk prompting aims to elicit world knowledge encoded in the model through a multi-step prompting scheme. The model is guided towards generating explicit clarification context for the original question that results in more faithful answers. Both, clarification and context are prompted to the model to predict the final answer.

An illustration of the self-talk prompting process is depicted in Fig~\ref{fig:selftalk}. 

Given a multiple-choice premise, the model is first prompted with the premise concatenated with one of multiple question prefixes. Prefixes are engineered for each dataset individually and depend on the type of posed questions. 
The language model completes several questions for each prefix, which are then used for extracting clarification answers. For each well-formed question multiple answers are generated through additional sampling. The prompt for the final answer generation consists of the original premise and the generated clarification.

\section{Visual Reasoning Benchmark Datasets}
\label{appendix:sec:datasets}
Here, we provide further details on the composition of the datasets used to evaluate visual reasoning based on VQA tasks.

The VQA-X dataset extends the COCO based VQA-v1~\cite{zitnick2016measuring} and v2~\cite{goyal2017making} datasets with human-annotated explanations. Similarly, ACT-X provides explanations for human activities and builds on the MPII Human Pose~\cite{andriluka20142D} dataset. Therefore, ACT-X is not originally a VQA task as the datasets contains an answer in the performed activity but no question. Nonetheless, the intended open-end activity classification is entailed by the VQA task with a questions such as \textit{'What is the person doing'}. Therefore, we construct a ACT-X based VQA task using question prompts similar to the one stated above. 
Lastly, the CLEVR-X dataset provides synthetically generated explanations for the CLEVR~\cite{johnson2017clevr} dataset. Although automatically generated, the contained ground truth labels are of similar quality as human generated once since they are generated from underlying CLEVR scene graphs using templates with varying wording.  

We note that the VQA-X test split is not publicly available wherefore we randomly split the original validation set into a custom validation and test set.
\section{Threshold in Automatized Feedback} 
\label{appendix:sec:filter_threshold}
\begin{table}[ht]
    \centering
    \small
    \begin{tabular}{l c c c c}
    
    \textbf{Threshold} & \textbf{\# samples}   & \textbf{B-4} & \textbf{R-L} & \textbf{C} \\ \hline
     R-L $\geq$ 0.7 & 1207 & 15.38 & 40.32  & 48.88  \\
     R-L $\geq$ 0.8 & \phantom{1}234  & 14.99  & 40.12 & 46.65 \\
     R-L $\geq$ 0.9 & \phantom{11}80   & 14.03  & 38.21 & 42.17 \\
     
    \end{tabular}
    \caption{Comparison of training MAGMA on VQA-X with self-generated training data vs ground truth explanations. Scores are reported on the validation dataset. Bleu-4, Rouge-L \& CIDEr score in \% where higher is better. Naturally, the number of samples decreases with higher thresholds. Additionally, higher thresholds also lead to a decrease of explanation quality.}
    \label{tab:sampling_thresholds}
\end{table}

Here, we perform a brief ablation study on choosing the right threshold for filtering generated samples. As discussed previously we establish a lower bound of R-L $\geq$ 0.7 based on manual inspection of generated samples. 
As the model trained using R-L $\geq$ 0.7 is only slightly worse than the one utilizing the respective ground truth explanation, increasing the ROUGE-L threshold is unlikely to yield any benefits in performance. With increasing threshold, the number of \textit{fitting} explanations decreases rapidly, thus the trade-off between the number of samples and their closeness to the ground truth is non-benefitial. On the contrary, we argue that maintaining some variance with respect to ground truth alignment can result in a more robust training process. Tab.~\ref{tab:sampling_thresholds} shows the results on the VQA-X validation set of 3 experiments that only differ in the threshold used for choosing \textit{fitting} samples and thus the number of training explanations. The resulting explanation scores decrease with decreasing number of samples disregarding their closeness to the ground truth. Therefore, we used the threshold of R-L $\geq$ 0.7 for all experiments.
Nonetheless, this experiment also demonstrates that the model is able to generalize on small sample sizes. This demonstrates the approach to be applicable to applications where the initial number of \textit{fitting} explanations is low.

\section{Training on Self-Generated Samples}
\label{appendix:sec:self-training}
MAGMA performing reasonably well on zero-shot reasoning enables generating training samples from the model itself to improve already existing capabilities. 
Next to the above ablation study justifying the sampling threshold ROUGE-L $\geq 0.7$, 
we describe and provide further evidence for the chosen sampling and tuning setup.

\paragraph{Sampling and feedback}
Initial sampling on the training datasets of VQA-X and CLEVR-X produced \textit{fitting} explanations for roughly 4\% and 10\% of the training set, respectively. In the case of ACT-X, we received significantly fewer \textit{fitting} explanations from the baseline model ($\leq 1\%$).

The feedback provided by selecting \textit{fitting} explanations can be fed back into the model by performing finetuning on these explanations. We employed training as described in the paper on filtered, self-generated explanations. 
Next, we analyze the training behavior of one iteration of \illume~for reasoning by comparing training on self-generated samples to training on ground truth data.
\paragraph{Self-generated vs. ground truth samples}
\begin{table}[t]
    \centering
    \small
    \begin{tabular}{l c c c}
    
    \textbf{Training samples}    & \textbf{B-4}$\uparrow$ & \textbf{R-L}$\uparrow$ & \textbf{C}$\uparrow$ \\ \hline
     Self-generated (w/ R-L $\geq$ 0.7) & 15.38 & 40.32 & 48.88 \\
     GT explanation (same as above)     & 17.80 & 41.31 & 50.68 \\
     Random GT samples                  & 19.93 & 44.28 & 64.50 \\
     
    \end{tabular}
    \caption{Comparison of training MAGMA on VQA-X with self-generated training data vs. ground truth (GT) explanations. Scores are reported on the validation set. Bleu-4, Rouge-L \& CIDEr score where higher is better. Training with self-generated explanations results in similar performance as training on the same GT explanations. However, other randomly drawn but equally sized subsets can lead to better scores. 
    }
    \label{tab:self_training}
\end{table}
\begin{table}[t]
    \centering
    \small
    \def\arraystretch{1.2}\tabcolsep=1.4pt
    \begin{tabular}{ l c @{\hskip 7pt} c @{\hskip 7pt} c @{\hskip 4pt} }
   \textbf{Setup} & \textbf{B-4$\uparrow$} & \textbf{R-L$\uparrow$} & \textbf{C$\uparrow$}  \\ \hline
     \textit{MAGMA$_{\text{base}}$} &  3.30 & 22.44  & 17.08   \\ \hline
    
    \textbf{(1)} No VQA & 0.83 & 16.04 & 14.81 \\
    \textbf{(2)} No add. VQA & 3.25 & 21.62 & 25.39\\
    \textbf{(3)} With VQA (\illume) & \textbf{6.52} & \textbf{32.46} & \textbf{43.66} \\
    
    \end{tabular}
    \caption{Ablation study on using VQA task in fine-tuning in addition to explanations. All 3 methods were tuned for 2 iterations on the \mbox{ACT-X} \illume~training data. \label{tab:cl_ablation}}
\end{table}
Tab.~\ref{tab:self_training} shows how training on noisy, self-generated explanations compares to training on ground truth data. We ran three training setups on VQA-X with the same hyperparameters and number of training samples that only differ in how training data is sourced. The first experiment uses 1207 \textit{fitting} explanations generated by the model. The second configuration considers the same 1207 samples as the previous one. However, it trains on the GT explanation instead. At the same time, the third setup randomly samples 1207 GT explanations.\footnote{Contrary to the iterative \illume~tuning process, we did not include any additional VQA samples for this experiment.}

The results for training on self-generated samples instead of the same GT explanations are slightly favorable towards using ground truth data. Presumably, this is due to the GT explanations in the training set being closer in distribution to the GT explanations of the validation set used for evaluation. Nonetheless, the difference remains negligible to the extent that it might even be attributed to noise in the training and evaluation loop. However, the subset of samples for which the VLM generates \textit{fitting} explanations appears not to be the optimal one concerning the generalization learned during training. 
Note that we drew multiple random samples of the same number of GT explanations from the training data, all of which resulted in a slightly better model than the sampled GT subset.

\begin{table*}[t]
    \small
    \def\arraystretch{1}\tabcolsep=2.5pt
    \centering
    \begin{tabular}{l c c c c | c c c c | c c c c}
    
                        &  \multicolumn{4}{c}{\textbf{VQA-X}}   &  \multicolumn{4}{c}{\textbf{ACT-X}}   &  \multicolumn{4}{c}{\textbf{CLEVR-X}} \\ 
     \textbf{Model}     & \textbf{B-4}$\uparrow$ & \textbf{R-L}$\uparrow$  &  \textbf{M}$\uparrow$  & \textbf{C}$\uparrow$ & \textbf{B-4}$\uparrow$ & \textbf{R-L}$\uparrow$     & \textbf{M}$\uparrow$ &   \textbf{C}$\uparrow$       & \textbf{B-4}$\uparrow$ & \textbf{R-L}$\uparrow$   & \textbf{M}$\uparrow$  &   \textbf{C}$\uparrow$  \\ \hline
     MAGMA      & 24.7  & 47.7 & 23.4 & 81.8  & 20.3  & 41.4   & 20.3 & 103.0  & 95.3  & 97.1  & 68.0 & 273.2 \\
     
    \end{tabular}
    \caption{Evaluation on the respective validation datasets of MAGMA explanation generated after finetuning. Scores refer to Bleu-4, Rouge-L \& CIDEr (\%) where higher scores are better. \label{tab:magma_tuned_val}}
   
\end{table*}
\begin{table*}[t]
    \small
    \def\arraystretch{1}\tabcolsep=2.5pt
    \centering
    \begin{tabular}{l c c c c | c c  c c | c c c  c}
    
                        &  \multicolumn{4}{c}{\textbf{VQA-X}}   &  \multicolumn{4}{c}{\textbf{ACT-X}}   &  \multicolumn{4}{c}{\textbf{CLEVR-X}} \\ 
     \textbf{Model}     & \textbf{B-4} & \textbf{R-L}  & \textbf{M}$\uparrow$ &   \textbf{C}          & \textbf{B-4} & \textbf{R-L}  & \textbf{M}$\uparrow$  &   \textbf{C}       & \textbf{B-4} & \textbf{R-L}  & \textbf{M}$\uparrow$  &   \textbf{C}  \\ \hline
     PJ-X  & 19.8  & 44.0  & 18.6 & 73.4 & 24.5  & 46.9  & 21.5 & 58.7 & 87.4 & 93.4    & 58.9 & 639.8    \\
     FM    & 24.4  & 47.4  &  19.5 & 88.8 & -     &-      & -   & -    & 78.8 & 85.8   & 52.5 & 566.8    \\
     MAGMA & 21.5  & 46.8  & 23.4 & 73.8 & 15.4  & 40.3   & 20.3 & 93.0 & 95.0 & 96.9   &  68.0 & 273.4 \\
     
    \end{tabular}
    \caption{Comparison of MAGMA tuned for explanation generation against state-of-the-art models on respective test sets. Scores refer to Bleu-4, Rouge-L \& CIDEr (\%) where higher scores are better. MAGMA remains competitive with the much more specialized architectures of PJ-X and FM.\label{tab:magma_tuned_comp}}
    
\end{table*}

\section{\illume~as Continual Learning Setup}\label{appendix:sec:continual_learning}
The iterative nature of fine-tuning in \illume~makes this in essence a continual learning task. Consequently, we have to address well known issues such as catastrophic forgetting \cite{mccloskey2989catastrophic} when training for more than one iteration. One of the measures employed by \illume~to tackle this problem is the extension of the training task and data to include VQA samples. Here we present a small ablation study providing further empirical evidence in favor of this technique. We fine-tuned Magma for two iterations on the filtered, self-generated samples from \illume~ and compare three setups. 

(1) which is trained solely on the explanation generation task without including the VQA task, (2) where we included the VQA task for each explanation sample, and (3) for which we added additional VQA samples as described for \illume.
The results are depicted in Tab. \ref{tab:cl_ablation}. It is apparent that after only two iterations the \illume~setup clearly outperforms the other approaches. The one with no additional VQA samples (2) barely improves over the baseline. The one trained on explanations only (2) already overfitted and achieves worse scores than zero-shot MAGMA. 
This effect would even reinforce itself on subsequent iterations with sampling being carried out on worse performing models. 
Consequently, we use the setup with additional VQA samples for \illume.

\section{Benchmarking MAGMA for VQA Explanation Generation}
\label{appendix:sec:gtexp}
In the main text of this work we evaluate \illume~on MAGMA. However, MAGMA also proves to perform excellent on visual reasoning using the common setting of learning from ground truth explanations.

To demonstrate MAGMA's performance, we provide a benchmark on the VQA-X, ACT-X and CLEVR-X datasets using ground truth data, and, importantly, compare the fine-tuned MAGMA models to current state-of-the-art (SOTA) architectures.
To the best of our knowledge there exist two models for the task of VQA explanation generation which are PJ-X  \cite{park2018multimodal} and FM  \cite{Wu2019faithful}.

We establish a baseline for MAGMAs potential performance on these datasets by fine-tune the model for each of them using the ground truth explanations of the entire training split using the previously described training setup.

Our goal is to provide an estimate of MAGMAs performance and not report new state-of-the-art results for relevant benchmarks. Therefore, we limited our hyperparamter tuning to a limited guided search of the hyperparamter space and did not perform any exhaustive grid searches. Consequently, the following results are solely intended to provide some intuition about MAGMAs performance within the limitations of evaluation with automated metrics. 

Tab.~\ref{tab:magma_tuned_val} shows the performance  of MAGMA on the respective validation datasets fine-tuned for each of the VQA-X, ACT-X and CLEVR-X. Tab.~\ref{tab:magma_tuned_comp} compares MAGMAs performance to the current state-of-the-art models PJ-X and LM on the respective test splits. 
We sourced the scores of both models from the respective papers.
As we are having to use custom validation and test splits, this results in us reporting scores for VQA-X on different splits than PJ-X and FM limiting the significance of these comparisons. Nonetheless, MAGMA remains competitive with both models across all three datasets. These results demonstrate the potential power of large-scale multimodal language models such as MAGMA, given that it is a general-purpose model for text generation on any multimodal inputs. Whereas PJ-X and FM were specifically designed to perform only this one task. 

Here, we investigate the influence of increasing MAGMA's capacity (as number of parameters) on the zero-shot performance of VQA explanation generation. Since the underlying language model of MAGMA extended only slightly outperforms the LM of the base model on natural language reasoning, we limit this comparison to VQA-X. 

\begin{table}[ht]
    \centering
    \def\arraystretch{1}\tabcolsep=7.pt
    \small
    \begin{tabular}{l l c c c c}
    \textbf{a)} & \textbf{Iteration} & \textbf{B-4} & \textbf{R-L} & \textbf{C}  & \textbf{\# Samples}  \\ \cline{2-6}
     &It 0 &  25.35 & 59.44 & 15.24 & $\phantom{111}\textrm{0}^{ \:\phantom{1}0.0\%}$  \\ 
     &It 1 & 12.12  & 32.95  & 20.59 & $\textrm{3342}^{ \:11.1\%}$\\
     &It 2 & \phantom{1}7.04  & 29.57  & 14.86 & $\textrm{1642}^{ \:\phantom{1}5.4\%}$ \vspace{1em} \\ 
    \textbf{b)}&\textbf{Iteration} & \textbf{B-4} & \textbf{R-L} & \textbf{C}  & \textbf{\# Samples}  \\ \cline{2-6}
     &It 0 &  14.27 & 58.93 & \phantom{1}6.56 & $\phantom{111}\textrm{0}^{ \:\phantom{1}0.0\%}$  \\
     &It 1 & \phantom{1}7.28  & 30.85  & 16.61 & $\textrm{3342}^{ \:11.1\%}$\\
     &It 2 & \phantom{1}6.25  & 28.77  & 15.27 & $\textrm{1642}^{ \:\phantom{1}5.4\%}$\\ 
    \end{tabular}
    \caption{Results of two iteration of \illume~on CLEVR-X. Bleu-4, Rouge-L \& CIDEr score reported on the a) validation split and b) training split.}
    \label{tab:st_clevrx}
\end{table}
%
     
%
     

\section{\illume~-- Extended Metrics}
\label{appendix:sec:extmetrics}
In Tab.~\ref{appendix:tab:exp_selftaught} we provide our \illume~results for VQA-X and ACT-X extended with further metrics. 
\begin{table*}[t]
    \centering
    \small
    \def\arraystretch{1}\tabcolsep=4.pt
    \begin{tabular}{l c c c c c c c c c}
    \multicolumn{9}{c}{\textbf{VQA-X}} \vspace{0.25em}\\
    \textbf{Iteration} & \textbf{B-1}$\uparrow$ $\Delta$ & \textbf{B-2}$\uparrow$ $\Delta$ & \textbf{B-3}$\uparrow$ $\Delta$ & \textbf{B-4}$\uparrow$ $\Delta$ & \textbf{R-L}$\uparrow$ $\Delta$ & \textbf{M}$\uparrow$ $\Delta$ & \textbf{C}$\uparrow$ $\Delta$  & \textbf{RV}  \\ \hline
     MAGMA$_{base}$ & 33.50 \phantom{$\pm$ 0.0} & 22.35 \phantom{$\pm$0.0} & 14.15\phantom{$\pm$ 0.0} & \phantom{1}9.16 \phantom{$-$0.0} & 32.45 \phantom{$-$0.0} & 12.27 \phantom{$\pm$ 0.0} & 31.08 \phantom{$-$0.0} & \phantom{1}0.0\% (0)\phantom{111}  \\ 
     It 1 & 42.54 $-$3.3 & 30.44 $-$1.8 & 20.73 $-$0.6 & 14.06 $+$\textbf{0.2} & 39.52 $-$0.7  & 14.96 $\pm$\textbf{0.0} &  44.57 $+$\textbf{3.4} &   \phantom{1}4.1\% (1207) \\
     It 2 & 44.54 $-$3.7 & 32.84 $-$3.7  & 23.17 $-$3.7  & 16.14 $-$3.5   & 41.03 $-$2.4   & 15.49 $-$3.0  &  48.62 $-$13.2 &   \phantom{1}6.6\% (1959)\\
     It 3 & 45.95 $-$3.3 & 34.17 $-$2.4  & 24.78 $-$1.4  & 17.42 $-$1.2  & 42.49 $+$\textbf{0.1}   & 16.06 $-$2.5  & 52.91 $-$5.2  &   \phantom{1}8.3\% (2446)\\
     It 4 & 47.12 $-$4.3  & 35.27 $-$2.4  & 25.33 $-$1.4  & 18.12 $-$1.1   & 42.83 $+$\textbf{0.1}   & 16.47 $-$2.2  &  57.11 $-$0.9  &   \phantom{1}9.3\% (2734)\\
     It 5 & 46.80 $-$5.8  & 35.57 $-$3.5  & 26.24 $-$1.9  & 19.35 $-$0.8   & 43.67 $-$0.2   & 16.64 $-$2.4  & 59.51 $-$0.7  &  10.1\% (2974)\\
     It 6 & 48.15 $-$5.5  & 36.45 $-$3.5  & 26.65 $-$2.4  & 19.54 $-$1.7   & 43.68 $+$\textbf{0.3}   & 16.87 $-$2.5  &  61.13 $-$3.9  &   10.9\% (3226)\\
     It 7 & 49.73 $-$3.0  & 37.85 $-$1.5  & 27.87 $-$0.5  & 20.13 $-$0.5   & 44.55 $+$\textbf{1.2}   & 17.46 $-$2.4  & 62.85 $+$\textbf{1.2}  &  11.5\% (3385)\\
     It 8 & 51.68 $-$1.1  & 38.97 $+$\textbf{0.3}  & 28.38 $+$\textbf{0.7}  & 20.56 $+$\textbf{0.7}   & 44.75 $+$\textbf{1.2}   & 17.52 $-$2.3  &  65.20 $+$\textbf{1.8}  &  12.0\% (3541)\\
     \hline 
     Test split (It 8) & 50.32 $-$1.4  & 37.20 $-$0.5  & 26.62 $-$0.1 & 19.01 $+$\textbf{0.3}  & 44.24 $+$\textbf{0.7}  & 16.52 $-$2.5  & 60.18 $+$2.7  &  12.0\% (3541)\\ 
     MAGMA$_{full}$ & 59.41 \phantom{$\pm$0.0} & 42.82 \phantom{$\pm$0.0} & 30.72 \phantom{$\pm$0.0} & 21.94 \phantom{$\pm$0.0} & 44.24 \phantom{$\pm$0.0} & 22.46 \phantom{$\pm$0.0} & 73.79 \phantom{$\pm$0.0} &  100.0\% (29459) \\
     \vspace{1em}
    \end{tabular}
   
    \begin{tabular}{l c c c c c c c c }
    \multicolumn{9}{c}{\textbf{ACT-X}} \vspace{0.25em}\\
    \textbf{Iteration} & \textbf{B-1}$\uparrow$ $\Delta$ & \textbf{B-2}$\uparrow$ $\Delta$ & \textbf{B-3}$\uparrow$ $\Delta$ & \textbf{B-4}$\uparrow$ $\Delta$ & \textbf{R-L}$\uparrow$ $\Delta$ & \textbf{M}$\uparrow$ $\Delta$ & \textbf{C}$\uparrow$ $\Delta$  & \textbf{RV}  \\ \hline
     MAGMA$_{base}$ & 31.83 \phantom{$-$00.0} & 13.99 \phantom{$-$00.0} & \phantom{1}6.72 \phantom{$-$0.0} & \phantom{1}3.30 \phantom{$-$0.0} & 22.44 \phantom{$-$0.0} & 12.23 \phantom{$-$0.0} & 17.08 \phantom{$-$00.0} & $-$  \\ \hline
     It 1 & 15.91 $-$18.0 & 10.31 $-$12.2 & \phantom{1}5.76 $-$8.8 & \phantom{1}3.63  $-$6.1 & 27.17 $-$7.2  & 11.47 $-$4.2  & 24.81  $-$26.3 &  \phantom{1}0.7\% (88)\phantom{11} \\
     It 2 & 25.99 $-$15.3 & 16.77 $-$11.3 & 10.13 $-$8.4 & \phantom{1}6.52 $-$6.3  & 32.46 $-$5.8 & 13.37  $-$5.0 & 43.66  $-$30.0 &   \phantom{1}4.0\% (503)\phantom{1} \\
     It 3 & 28.12 $-$13.1 & 18.38 $-$10.1 & 11.47 $-$7.8 & \phantom{1}7.58 $-$6.0  & 33.79 $-$5.4  & 14.39 $-$4.2 & 46.98 $-$29.0 &   \phantom{1}8.4\% (1054) \\
     It 4 & 30.51 $-$10.0 & 20.11 $-$\phantom{1}7.8 & 12.87 $-$6.0 & \phantom{1}8.79$-$4.7  & 34.55  $-$4.7  & 15.01 $-$3.6 & 52.02 $-$29.1 & 12.4\%  (1438)\\
     It 5 & 32.27 $-$\phantom{1}9.4 & 21.41 $-$\phantom{1}7.5 & 13.89 $-$5.9 & \phantom{1}9.54  $-$4.7  & 35.55  $-$4.1  & 15.54 $-$3.4 & 58.38 $-$24.0 &  16.2\% (1731) \\
     It 6 & 33.35 $-$\phantom{1}8.4 & 22.18 $-$\phantom{1}6.7 & 14.51 $-$5.1 & \phantom{1}9.99 $-$4.0  & 35.79 $-$3.8  & 16.14 $-$3.0 & 62.31 $-$20.2 &  17.8\% (1967) \\
     It 7 & 35.22 $-$\phantom{1}6.9 & 23.40 $-$\phantom{1}5.7 & 15.43 $-$4.2 & 10.66  $-$3.2  & 36.22  $-$2.9  & 16.48 $-$2.2 & 62.84  $-$19.7 &  19.6\% (2131) \\
     It 8 & 35.41 $-$\phantom{1}5.7 & 23.46 $-$\phantom{1}4.4 & 15.40 $-$3.1 & 10.64  $-$2.2  & 36.71  $-$1.3  & 16.48 $-$2.2 & 65.83  $-$11.1 & 20.4\% (2272) \\
     It 9 & 35.87 $-$\phantom{1}5.1 & 23.74 $-$\phantom{1}3.8 & 15.50 $-$2.7 & 10.65  $-$2.0  & 36.37  $-$1.1  & 16.66 $-$1.6 & 65.02  $-$\phantom{1}7.2 & 21.5\% (2387)\\
     \hline 
     Test split (It 9) & 35.30 $-$\phantom{1}5.5& 23.60 $-$\phantom{1}1.3 &15.57 $-$\phantom{1}3.2  & 10.79 $-$2.5   & 36.13  $-$1.9  & 16.53 $-$1.8 & 64.70  $-$13.2 &  21.5\% (2387)\\ 
     MAGMA$_{full}$ & 42.68 \phantom{$\pm$00.0} & 38.07 \phantom{$\pm$00.0} & 21.05 \phantom{$\pm$0.0} & 15.36 \phantom{$\pm$.0} & 40.34 \phantom{$\pm$0.0} & 19.56 \phantom{$\pm$0.0} & 92.96 \phantom{$\pm$0.0} &  100.0\% (29459) \\
     
    \end{tabular}
   \caption{Iterative process of \illume~on VQA-X (top) and ACT-X (bottom) until scores plateau on the validation set. 
    $\Delta$ values next to the scores indicate the difference between training on self-generated samples vs. the same amount of GT samples, with positive scores indicating that \illume~outperforms training on GT (\textbf{bold}) and vice versa.
    MAGMA$_{base}$ refers to zero-shot (\textit{It 0}) performance and MAGMA$_{full}$ refers to the model tuned on the entirety of the GT training set,.
    Additionally, RV displays the relative value wrt. total amount of samples in original training set. 
    The bottom rows show scores on the test set. 
    Bleu-1 through Bleu-4, Rouge-L, Meteor\& CIDEr scores are shown (higher is better).   \label{appendix:tab:exp_selftaught}}
\end{table*}
\section{Flaws in Logical Reasoning}
\label{appendix:sec:logical}
As mentioned in the main text, we frequently observed shortcoming of VLMs ability to generalize to logical reasoning. The following experiments provide further evidence in this regard.

Tab.~\ref{tab:st_clevrx}(a) shows the progress over two iterations of \illume~tuning on the CLEVR-X validation split. With each iteration of training the quality of textual explanations decreases instead of improving. This also results in fewer \textit{fitting} explanations being generated, exacerbating this effect further. Furthermore, finetuning on the 5-10\% subset of the training data used in self-talk fails to generalize explanations to the rest of the training set. Tab.~\ref{tab:st_clevrx}(b) indicates that all but the CIDEr score drop significantly after fine-tuning on the training split as well. 


\end{document}